\documentclass[runningheads]{llncs}
\usepackage{graphicx}
\usepackage{amsmath,amssymb} 
\usepackage{color}
\usepackage[width=122mm,left=12mm,paperwidth=146mm,height=193mm,top=12mm,paperheight=217mm]{geometry}
\usepackage{xcolor}
\usepackage{amsmath}
\usepackage{amssymb}
\usepackage{booktabs}
\usepackage{multirow}
\usepackage[utf8]{inputenc}
\usepackage{subcaption}
\definecolor{bblue}{HTML}{4F81BD}
\definecolor{rred}{HTML}{C0504D}
\definecolor{ggreen}{HTML}{9BBB59}
\definecolor{ppink}{HTML}{7A61CF}
\definecolor{ppurple}{HTML}{9F4C7C}
\usepackage{adjustbox}
\usepackage{array}
\usepackage{tikz,pgfplots}
\usepackage{dsfont}
\usepackage{hyperref}
\usepackage{pifont}

\definecolor{rosso}{RGB}{220,57,18}
\definecolor{giallo}{RGB}{255,153,0}
\definecolor{blu}{RGB}{102,140,217}
\definecolor{verde}{RGB}{16,150,24}
\definecolor{viola}{RGB}{153,0,153}

\makeatletter

\tikzstyle{chart}=[
    legend label/.style={font={\scriptsize},anchor=west,align=left},
    legend box/.style={rectangle, draw, minimum size=5pt},
    axis/.style={black,semithick,->},
    axis label/.style={anchor=east,font={\tiny}},
]

\tikzstyle{bar chart}=[
    chart,
    bar width/.code={
        \pgfmathparse{##1/2}
        \global\let\bar@w\pgfmathresult
    },
    bar/.style={very thick, draw=white},
    bar label/.style={font={\bf\small},anchor=north},
    bar value/.style={font={\footnotesize}},
    bar width=.75,
]

\tikzstyle{pie chart}=[
    chart,
    slice/.style={line cap=round, line join=round, very thick,draw=white},
    pie title/.style={font={\bf}},
    slice type/.style 2 args={
        ##1/.style={fill=##2},
        values of ##1/.style={}
    }
]

\pgfdeclarelayer{background}
\pgfdeclarelayer{foreground}
\pgfsetlayers{background,main,foreground}

\newcommand{\pie}[3][]{
    \begin{scope}[#1]
    \pgfmathsetmacro{\curA}{90}
    \pgfmathsetmacro{\r}{1}
    \def\c{(0,0)}
    \node[pie title] at (90:1.3) {#2};
    \foreach \v/\s in{#3}{
        \pgfmathsetmacro{\deltaA}{\v/100*360}
        \pgfmathsetmacro{\nextA}{\curA + \deltaA}
        \pgfmathsetmacro{\midA}{(\curA+\nextA)/2}

        \path[slice,\s] \c
            -- +(\curA:\r)
            arc (\curA:\nextA:\r)
            -- cycle;
        \pgfmathsetmacro{\d}{max((\deltaA * -(.5/50) + 1) , .5)}

        \begin{pgfonlayer}{foreground}
        \path \c -- node[pos=\d,pie values,values of \s]{$\v\%$} +(\midA:\r);
        \end{pgfonlayer}

        \global\let\curA\nextA
    }
    \end{scope}
}

\newcommand{\legend}[2][]{
    \begin{scope}[#1]
    \path
        \foreach \n/\s in {#2}
            {
                  ++(0,-10pt) node[\s,legend box] {} +(5pt,0) node[legend label] {\n}
            }
    ;
    \end{scope}
}

\usepackage{array,hhline}
\usepackage{pgf}

\newcommand*{\gray}{gray}

\newcommand*{\colorme}[1]{%
    \pgfmathparse{#1<.5?1:0}%
    \ifnum\pgfmathresult=0\relax\color{white}\fi
    \pgfmathparse{1-#1}
    \expandafter\cellcolor\expandafter[\expandafter\gray\expandafter]\expandafter{\pgfmathresult}%
    #1%
}
\newlength{\tabwidth}
\usepackage{adjustbox}
\usepackage{array}

\newcolumntype{R}[2]{%
    >{\adjustbox{angle=#1,lap=\width-(#2)}\bgroup}%
    l%
    <{\egroup}%
}

\newcommand{\cmark}{\ding{51}}
\newcommand{\xmark}{\ding{55}}

\def\ConvLSTM{ConvLSTM }

\begin{document}
\pagestyle{headings}
\mainmatter
\def\ECCV18SubNumber{2973}  
\title{Recurrent Neural Networks\\for Semantic Instance Segmentation}

\author{Anonymous ECCV submission}
\institute{Paper ID \ECCV18SubNumber}

\author{%
Amaia Salvador$^1$, Míriam Bellver$^2$, Víctor Campos$^2$, Manel Baradad$^1$ \\ Ferran Marques$^1$  Jordi Torres$^2$ \and Xavier Giro-i-Nieto$^1$ }

\institute{$^1$Universitat Politècnica de Catalunya
$^2$Barcelona Supercomputing Center}
\maketitle





\begin{abstract}
We present a recurrent model for semantic instance segmentation that sequentially generates binary masks and their associated class probabilities for every object in an image. Our proposed system is trainable end-to-end from an input image to a sequence of labeled masks and, compared to methods relying on object proposals, does not require post-processing steps on its output. We study the suitability of our recurrent model on three different instance segmentation benchmarks, namely Pascal VOC 2012, CVPPP Plant Leaf Segmentation and Cityscapes. Further, we analyze the object sorting patterns generated by our model and observe that it learns to follow a consistent pattern, which correlates with the activations learned in the encoder part of our network. 
\end{abstract}

\section{Introduction}
\label{sec:intro}



Semantic instance segmentation is defined as the task of assigning a binary mask and a categorical label to each object in an image. It is often understood as an extension of object detection where, instead of bounding boxes, accurate binary masks must be predicted. Current state of the art methods for semantic instance segmentation \cite{sds,hypercolumns,r2ios,fcis,mnc,maskrcnn} extend object detection pipelines based on object proposals \cite{faster-rcnn} by incorporating an additional module that is trained to generate a binary mask for each object proposal. Such architectures follow a two-stage procedure, i.e. a set of object-prominent proposal locations are selected first, and then each of them is given a score, a categorical label and a binary mask. Typically, the number of selected locations is much greater than the actual number of objects that appear in the image, meaning that post-processing is needed to select the subset of predictions that better covers all the objects and discard the rest. Although in most recent works the two different stages (i.e. proposal generation and scoring) are optimized jointly \cite{r2ios,fcis,mnc,maskrcnn}, the objective function still does not directly model the target task, but a surrogate one which is easier to handle at the cost of an additional filtering step.


Given enough training data and computational power, a great variety of automatic tasks such as object recognition \cite{krizhevsky2012_alexnet}, machine translation \cite{sutskever2014_seq2seq}, speech recognition \cite{graves2014towards} or self-driving cars \cite{bojarski2016_nvidiacar} have seen a boost of performance thanks to models trained end-to-end, 
i.e.~not imposing intermediate representations and directly learning to map the input to the desired output.
The novelty of our work is formulating and solving the semantic instance segmentation task end-to-end.

While most computer vision systems analyze images in a single step, the human exploration of static visual inputs is actually a sequential process \cite{porter2007,amor2016persistence} that involves reasoning about objects that compose the scene and their relationships. 
Inspired by this behavior, we design a model that performs a sequential analysis of the scene to deal with complex object distributions and make predictions that are coherent with each other. 
We take advantage of the capability of Recurrent Neural Networks 
to generate sequences out of a single input \cite{vinyals2015show,stewart2016end} and cast semantic instance segmentation as a sequence prediction task.
The model is trained to freely choose the scanpath over the image that maximizes the quality of the segmented instances, which allows us to conduct a detailed study about how it learns to explore images. The object discovery patterns we find are consistent and related to the relative layout of objects in the scene.

Recent works \cite{romeraparedes,renzemel} have also proposed sequential solutions for instance segmentation. These are, however, trained to produce a sequence of class-agnostic masks and must be either evaluated on single-class benchmarks or require a separate method to provide a categorical label for each predicted object. Both \cite{romeraparedes,renzemel} impose intermediate representations by using a pre-processed input consisting of a foreground/background mask and instance-level angle information \cite{renzemel} or using an encoder pre-trained for semantic instance segmentation \cite{romeraparedes}. Based on these works, we develop a true end-to-end recurrent system that provides a sequence of semantic instances as an output (i.e.~both binary masks and categorical labels for all objects in the image) \emph{directly} from image pixels. 

The contributions of this work are threefold: (a) we present the first end-to-end recurrent model for semantic instance segmentation, (b) we show its competitive performance against previous sequential methods on three instance segmentation benchmarks, and (c) we thoroughly analyze its behavior in terms of the object discovery patterns that it follows.

\section{Related Work}
\label{sec:relatedwork}
Most works on semantic instance segmentation inherit their foundations from object detection solutions, augmenting them to segment object proposals \cite{sds,hypercolumns} and adding post-processing stages to refine the predictions \cite{multi_instance}.
More recent works build on top of Faster R-CNN \cite{faster-rcnn} by adding a cascade of predictors \cite{mnc,dai_instance} and iterative refinement of masks \cite{r2ios}. 
In contrast with cascade-based methods \cite{mnc,r2ios,dai_instance}, He et al. \cite{maskrcnn} design an architecture that predicts bounding boxes, segments and class scores in parallel given the output of a fully convolutional network (hence, no chain reliance is imposed).
Other works have presented alternative methods to the proposal-based pipelines by treating the image holistically. These include combining object detection and semantic segmentation pipelines with Conditional Random Fields \cite{pixelwise}, learning a watershed transform on top of a semantic segmentation \cite{watershed_urtasun} or clustering object pixels with metric learning \cite{vangool}.



Our model is closer to recent works that formulate the problem of instance segmentation with sequential methods, which predict different object instances one at a time. Ren \& Zemel \cite{renzemel} propose a complex multi-task pipeline for instance segmentation that predicts the box coordinates for a different object at each time step using recurrent attention. These bounding boxes are then used to select the image location and predict a binary mask for the object. Their model uses an additional input consisting of a canvas that is composed of the union of the binary masks that have been previously predicted. This architecture resembles two-stage proposal-based ones \cite{sds,r2ios,maskrcnn} in the sense that it is also composed of two separate modules, one predicting location coordinates and one to produce a binary mask within this location. The main difference between these works and \cite{renzemel} is that objects are predicted one at a time and are dependent on each other. Romera-Paredes \&  Torr \cite{romeraparedes} choose to use a recurrent decoder that stores information about previously found objects in its hidden state. Their model is composed of Convolutional LSTMs \cite{ConvLSTM} that receive features from a pretrained model for semantic segmentation \cite{fcn} and outputs the separate object segments for the image. 


While proposal-based methods have shown impressive performance, they generate an excessive number of predictions and rely on an external post-processing step for filtering them out, e.g.~non-maximum suppression. Our proposed recurrent model optimizes an objective which better matches the conditions at inference time, as it is trained to predict the final semantic instance segmentation directly from image pixels.
All previous sequential methods \cite{romeraparedes,renzemel} are class-agnostic and, although \cite{renzemel} reports results for semantic instance segmentation benchmarks, class probabilities for their predicted segments are obtained from the output of a separate model trained for semantic segmentation. To the best of our knowledge, our proposed method is the first to directly tackle semantic instance segmentation with a fully end-to-end recurrent approach that maps image pixels to a variable length sequence of objects represented with binary masks and categorical labels.



\section{Model}
\label{sec:model}




Given an input image $x$, the goal of semantic instance segmentation is to provide a set of masks and their corresponding class labels, $y = \{ y_1, \dotsc, y_n \}$. The cardinality of the output set, i.e.~the number of instances, depends on the input image and thus the model needs to be able to handle variable length outputs. This poses a challenge for feedforward architectures, which emit outputs of fixed size. Similarly to previous works involving sets \cite{vinyals2015pointer,vinyals2016matching,romeraparedes}, we propose a recurrent architecture that outputs a sequence of masks and labels, $\hat{y} = ( \hat{y}_1, \dotsc, \hat{y}_{\hat{n}} )$. At any given time step $t \in \{1,\dotsc,\hat{n}\}$, the prediction is of the form $\hat{y}_t = \{\hat{y}_{m},\hat{y}_{b}, \hat{y}_{c},\hat{y}_{s}\}$, where $\hat{y}_{m} \in [0,1]^{h \times w}$ is the binary mask, $\hat{y}_{b}\in [0,1]^{4}$ are the bounding box coordinates 
normalized by the image dimensions, $\hat{y}_{c}\in [0,1]^{C}$ are the probabilities for the $C$ different categories, and $\hat{y}_{s}\in [0,1]$ represents the objectness score, which is the stopping criterion at test time. Obtaining bounding box annotations from the segmentation masks is straightforward and it adds an additional training signal, which resulted in better performing models in our experiments.

We design an encoder-decoder architecture that resembles typical ones from semantic segmentation works \cite{fcn,unet}, where skip connections from the layers in the encoder are used to recover low level features that are helpful to obtain accurate segmentation outputs. The main difference between these works and ours is that our decoder is recurrent, enabling the prediction of one instance at a time instead of a single semantic segmentation map where all objects are present, thus allowing to naturally handle variable length outputs. 


\subsection{Encoder}

We use a ResNet-101 \cite{resnet} model pretrained on ImageNet \cite{imagenet} for image classification as an encoder. We truncate the network at the last convolutional layer, thus removing the last pooling layer and the final classification layer. The encoder takes an RGB image $x\in \mathbb{R}^{h \times w \times 3}$ and extracts features from the different convolutional blocks of the base network $ F = \mathrm{encoder}(x)$. $F$ contains the output of each block $F =[f_{0},f_{1},f_{2},f_{3},f_{4}]$, where $f_{0}$ corresponds to the output of the deepest block, and $f_{4}$ is the output of the block whose input is the image (i.e.~$f_{4\ldots0}$ correspond to the output of $\text{ResBlock}_{1\ldots5}$ in ResNet-101, respectively).

\begin{figure}
\vspace*{-\baselineskip}
  \centering
  \includegraphics[width=\textwidth]{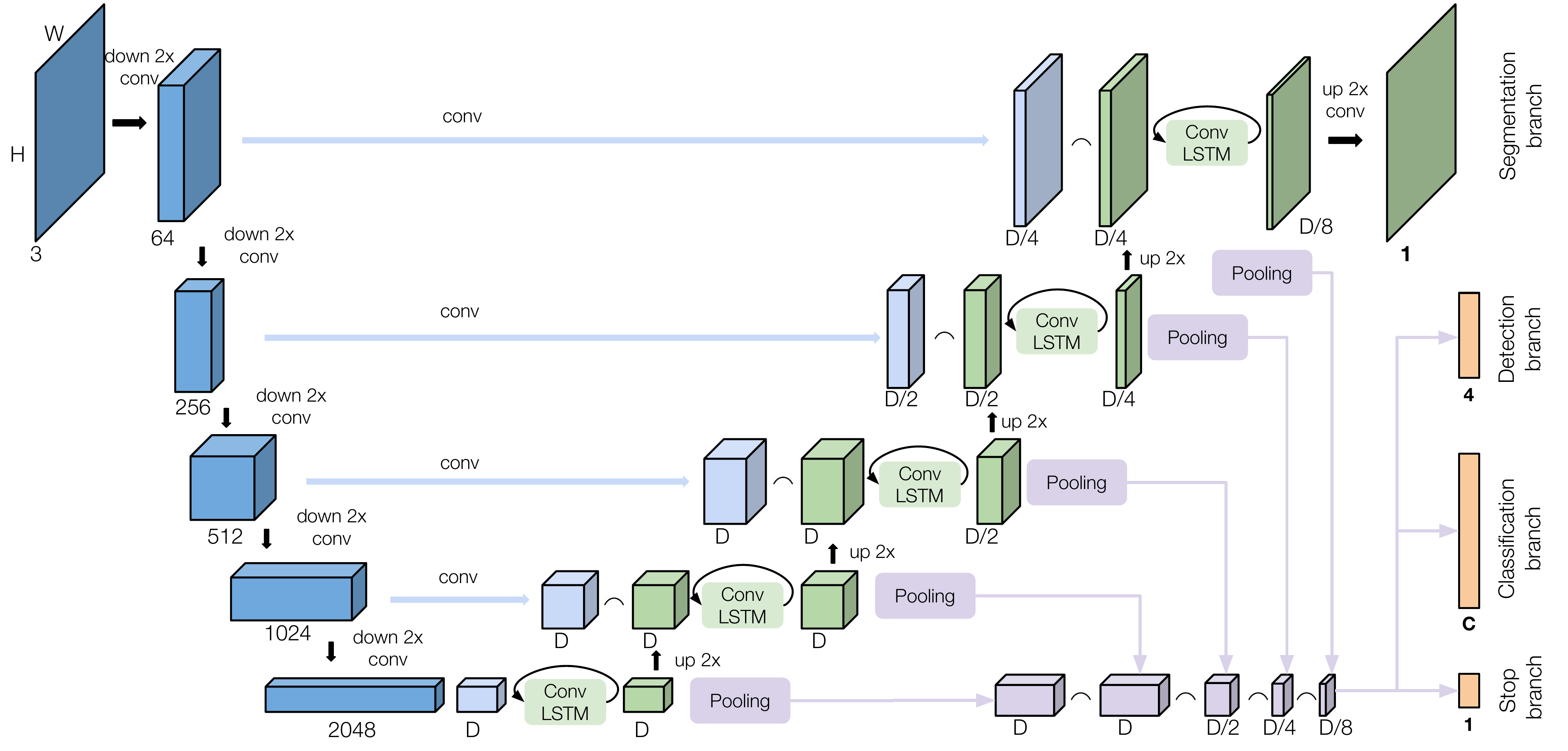}
  \caption{Our proposed recurrent architecture for semantic instance segmentation.}
  \label{fig:full_architecture}
\vspace*{-\baselineskip}
\end{figure}

\vspace*{-\baselineskip}
\subsection{Decoder}


The decoder receives as input the convolutional features $F$ and outputs a set of 
$\hat{n}$ 
predictions, being 
$\hat{n}$
variable for each input image. Similarly to \cite{romeraparedes}, we use the Convolutional LSTMs \cite{ConvLSTM} as the basic block of our decoder, in order to naturally handle 3-dimensional convolutional features as input and preserve spatial information. While \cite{romeraparedes} uses a two-layer Convolutional LSTM module that receives the output of the last layer of their encoder, we design a hierarchical recurrent architecture that can leverage features from the encoder at different abstraction levels. We design an upsampling network composed of a series of \ConvLSTM layers, whose outputs are subsequently merged with the side outputs $F$ from the encoder. This merging can be seen as a form of skip connection that bypasses the previous recurrent layers. Such architecture allows the decoder to reuse low level features from the encoder to refine the final segmentation. Additionally, since we are using a recurrent decoder, the reliance on these features can change across different time steps. 





The output of the $i^{th}$ ConvLSTM layer in time step $t$, $h_{i,t}$, depends on both (a) the input it receives from the encoder and its preceding ConvLSTM layer and (b) its hidden state representation in the previous time step $h_{i,t-1}$: 
\begin{equation}
\label{eq:ith_convlstm}
h_{i,t} = \mathrm{ConvLSTM_{i}}(\;[\; B_{2}(h_{i-1,t}) \; | \;S_{i} \;] , h_{i,t-1} \;)
\end{equation}
where $B_{2}$ is the bilinear upsampling operator by a factor of 2, $h_{i-1,t}$ is the hidden state of the previous \ConvLSTM layer and $S_{i}$ is the result of projecting $f_{i}$ to have lower dimensionality via a convolutional layer.

Equation \ref{eq:ith_convlstm} is applied in chain for $i \in \{1,\dotsc,n_b\}$, being $n_b$ the number of convolutional blocks in the encoder ($n_b=5$ in ResNet). $h_{0,t}$ is obtained by a \ConvLSTM with $S_{0}$ as input (i.e.~no skip connection):

\begin{equation}
\label{eq:ht0}
h_{0,t} = \mathrm{ConvLSTM_{0}}(S_{0}, h_{0,t-1})
\end{equation}

We set the first two \ConvLSTM layers to have dimension $D$, and set the dimension of the remaining ones to be the one in the previous layer divided by a factor of 2. All \ConvLSTM layers use $3 \times 3$ kernels which, compared to $1 \times 1$ ConvLSTM units used in \cite{romeraparedes}, have a larger receptive field which can model instances that are far apart more easily. Finally, a single-kernel $1 \times 1$ convolutional layer with sigmoid activation is used to obtain a binary mask of the same resolution as the input image.

The bounding box, class and stop prediction branches consist of three separate fully connected layers to predict the 4 box coordinates, the category of the segmented object and the objectness score at time step $t$. These three layers receive the same input $h_{t}$, which is obtained by concatenating the max-pooled hidden states of all \ConvLSTM layers in the network. Figure \ref{fig:full_architecture} shows the details of the recurrent decoder for a single time step.


\subsection{Training}

The parameters of our model are estimated by optimizing a multi-task objective composed of four different terms:



\noindent{\textbf{Segmentation loss ($\mathbf{L_m}$):}} similarly to other works \cite{romeraparedes,renzemel}, we use the soft intersection over union loss (sIoU) as the cost function between the predicted mask $\hat{y}$ and the ground truth mask $y$,
$\mathrm{sIoU}(\hat{y},y) = 1 - \frac{\left\langle \hat{y},y \right\rangle}{\left\lVert \hat{y} \right\rVert_1 + \left\lVert y \right\rVert_1 - \left\langle \hat{y},y \right\rangle}$.




We do not impose any specific instance order to match the predictions of our model with the objects in the ground truth. Instead, we let the model decide which output permutation is the best and sort the ground truth accordingly\footnote{We also experimented with forcing the output sequence to follow hand-designed patterns, but it resulted in low-performing models.}. We assign a prediction to each of the ground truth masks by means of the Hungarian algorithm, using sIoU as the cost function. Given a sequence of predicted masks $\hat{y}_{m} = (\hat{y}_{m,1},\dotsc,\hat{y}_{m,\hat{n}})$ and the set of ground truth masks $y_{m} = \{y_{m,1},\dotsc,y_{m,n}\}$, the segmentation loss $L_{m}$ can be expressed as:

\begin{equation}
L_{m}(\hat{y}_{m},y_{m},\delta) = \sum\limits_{t=1}^{\hat{n}}\sum\limits_{t'=1}^{n} sIoU(\hat{y}_{m,t},y_{m,t'}) \delta_{t,t'} \\
\end{equation}
where $\delta$ is the matrix of assignments. $\delta_{t,t'}$ is 1 when the predicted and ground truth masks $\hat{y}_{m,t}$ and $y_{m,t'}$ are matched and 0 otherwise. In the case where $\hat{n} > n$, gradients for predictions at $t > n$ are ignored.


\noindent{\textbf{Classification loss ($\mathbf{L_c}$):}} our network outputs class probabilities for each of the predicted masks. Given the sequence of class probabilities  $\hat{y}_{c} = (\hat{y}_{c,1},\dotsc,\hat{y}_{c,\hat{n}})$ and the set of ground truth one-hot class vectors $y_{c} = \{y_{c,1},\dotsc,y_{c,n}\}$, the classification loss is computed as the categorical cross entropy between the matched pairs determined by $\delta$.



\noindent{\textbf{Detection loss ($\mathbf{L_b}$):}} given the sequence of predicted bounding box coordinates  $\hat{y}_{b} = (\hat{y}_{b,1},\dotsc,\hat{y}_{b,\hat{n}})$ and the ground truth $y_{b} = \{y_{b,1},\dotsc,y_{b,n}\}$, the penalty term $L_{b}$ for bounding box regression is given by the mean squared error between the box coordinates of matched pairs determined by $\delta$.

\noindent{\textbf{Stop loss ($\mathbf{L_s}$):}} the model emits an objectness score at each time step, $\hat{y}_{s,t}$. It is optimized with a loss term defined as the binary cross entropy between $\hat{y}_{s,t}$ and $\mathds{1}_{t \leq n}$, where $n$ is the number of instances in the image.


The total loss is the weighted sum of the four terms: $L_{m} + \alpha L_{b} + \lambda L_{c} + \gamma L_{s}$, where loss terms are subsequently added as training progresses. When training for datasets with a high number of objects per image (i.e.~Cityscapes and CVPPP) we use curriculum learning \cite{curriculum} to guide the optimization process, where we begin optimizing the model to predict only two objects and increase this value by one once the validation loss plateaus.


\section{Experiments}
\label{sec:experiments}


Experiments are implemented with PyTorch\footnote{\url{http://pytorch.org/}}. Code and models will be publicly released upon acceptance. The choice of hyperparameters and other training details for each dataset are provided in the supplementary material.

\subsection{Datasets and metrics}
We evaluate our models on three benchmarks previously used for semantic instance segmentation that differ from each other in terms of the average amount of objects per image. This diversity in datasets will allow assessing our model based on the length of the sequence to be generated.


\noindent \textbf{Pascal VOC 2012 \cite{PASCAL_VOC}} contains objects of 20 different categories and an average of 2.3 objects per image. Despite having a small number of objects on average, images in this dataset are complex and substantially different from each other in terms of the objects spatial arrangement, scale and pose. Following standard practices in \cite{r2ios,vangool,mpa}, we train with the additional annotations from \cite{boundaries} and evaluate on the original validation set, composed of 1,449 images. 

\noindent \textbf{CVPPP Plant Leaf Segmentation \cite{cvppp}} is a small dataset of images of different plants. We follow the same scheme as in \cite{romeraparedes,renzemel}, using 
only 128 images from the A1 subset for training. The number of leaves per image ranges from 11 to 20, with an average of 16.2. Results are evaluated on 33 test images. While the number of objects per image is significantly higher than in Pascal VOC, this dataset only contains objects from a single category and images present structural similarities that facilitate the task.

\noindent \textbf{Cityscapes \cite{cityscapes}} contains 5,000 street-view images containing objects of 8 different categories. The dataset is split in 2,975 images for training, 500 for validation and 1,525 for testing. There are, on average, 17.5 objects per image in the training set, with the number of objects ranging from 0 to 120. The large number of instances per image makes this dataset particularly challenging for our model. 

We resize images to $256 \times 256$ pixels for Pascal VOC, $256 \times 512$ for Cityscapes and $500 \times 500$ for CVPPP. We evaluate the CVPPP dataset with the symmetric best dice (SBD) and the difference in count (DiC) as in \cite{cvppp}. For Cityscapes and Pascal VOC we report the average precision $AP$ at different IoU thresholds.

\begin{table}[]
\centering
\begin{tabular}{@{}lccccccccc@{}}
\toprule
\multicolumn{1}{c}{} 			& \textbf{Rec} 		& \textbf{Cls} & \textbf{Pascal VOC}  & \multicolumn{2}{c}{\textbf{CVPPP}} & \multicolumn{4}{c}{\textbf{Cityscapes}}\\ 
                           		&           &          & $AP_{person,50}$     & SBD  $\uparrow$       & DiC $\downarrow$     & AP  & $AP_{50}$  & $AP_{car}$  & $AP_{car,50}$         \\  \midrule

\cite{renzemel}       			&  \xmark         & \xmark         &     $-$    &      $\mathbf{84.9( \pm 4.8)} $      &   $\mathbf{0.8( \pm 1.0)}$             &  $\mathbf{9.5}$   &      $\mathbf{18.9}$ &    $\mathbf{27.5}$     &      $41.9$            \\
\cite{romeraparedes}       		& \cmark         & \xmark        &        $46.6$          &        $56.8(\pm 8.2)$     &    $1.1( \pm 0.9)$ &           $-$  &  $-$     &   $ - $    &     $-$            \\
\cite{romeraparedes}  + CRF  	& \cmark         & \xmark        &    $50.1$      &        $66.6( \pm 8.7) $    &   $1.1( \pm 0.9)  $          &  $-$   &   $ -$   &   $ - $    &    $ - $            \\
Ours                 			& \cmark         & \cmark        &      $\mathbf{60.7}$      &    $74.7( \pm 5.9)$         &       $1.1( \pm 0.9)$        &       $7.8$   &   $17.0$    &   $25.8$      &    $\mathbf{45.7}$         \\ \bottomrule
\end{tabular}
\caption{Comparison against state of the art sequential methods for semantic instance segmentation. We specify whether the method is recurrent (Rec) and produces categorical probabilities (Cls).}
\label{tab:seq_sota}
\vspace*{-\baselineskip}
\end{table}

\vspace*{-\baselineskip}
\subsection{Comparison with sequential methods}

We compare our results against other sequential models for instance segmentation \cite{romeraparedes,renzemel}. Table \ref{tab:seq_sota} summarizes the results. 

\begin{figure}
\vspace*{-\baselineskip}
\begin{subfigure}[t]{0.5\textwidth}
  \includegraphics[width=\columnwidth]{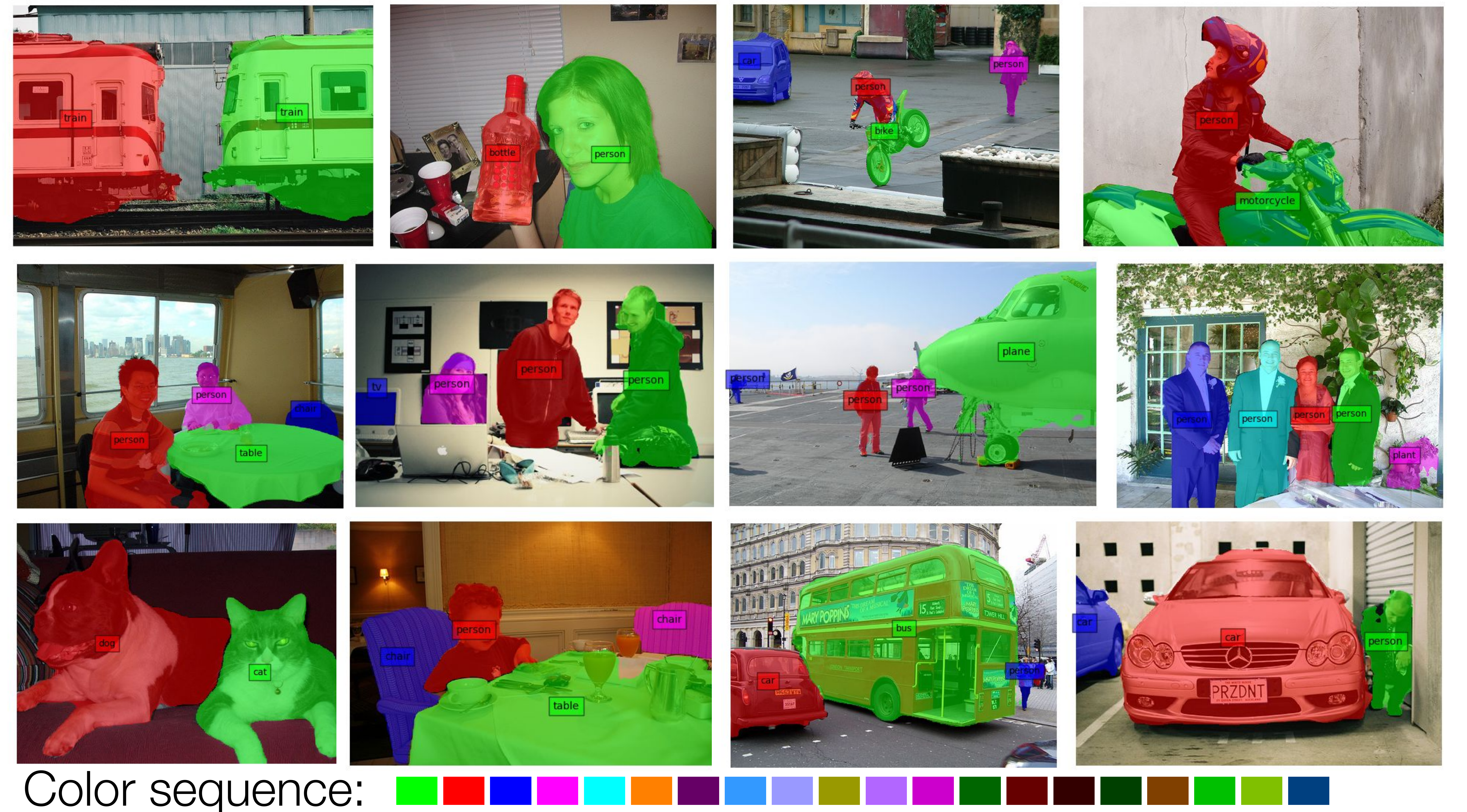}
  \caption{Pascal VOC 2012}
  \label{fig:pascal_examples}
\end{subfigure}
\begin{subfigure}[t]{0.5\textwidth}
  \centering
  \includegraphics[width=\columnwidth]{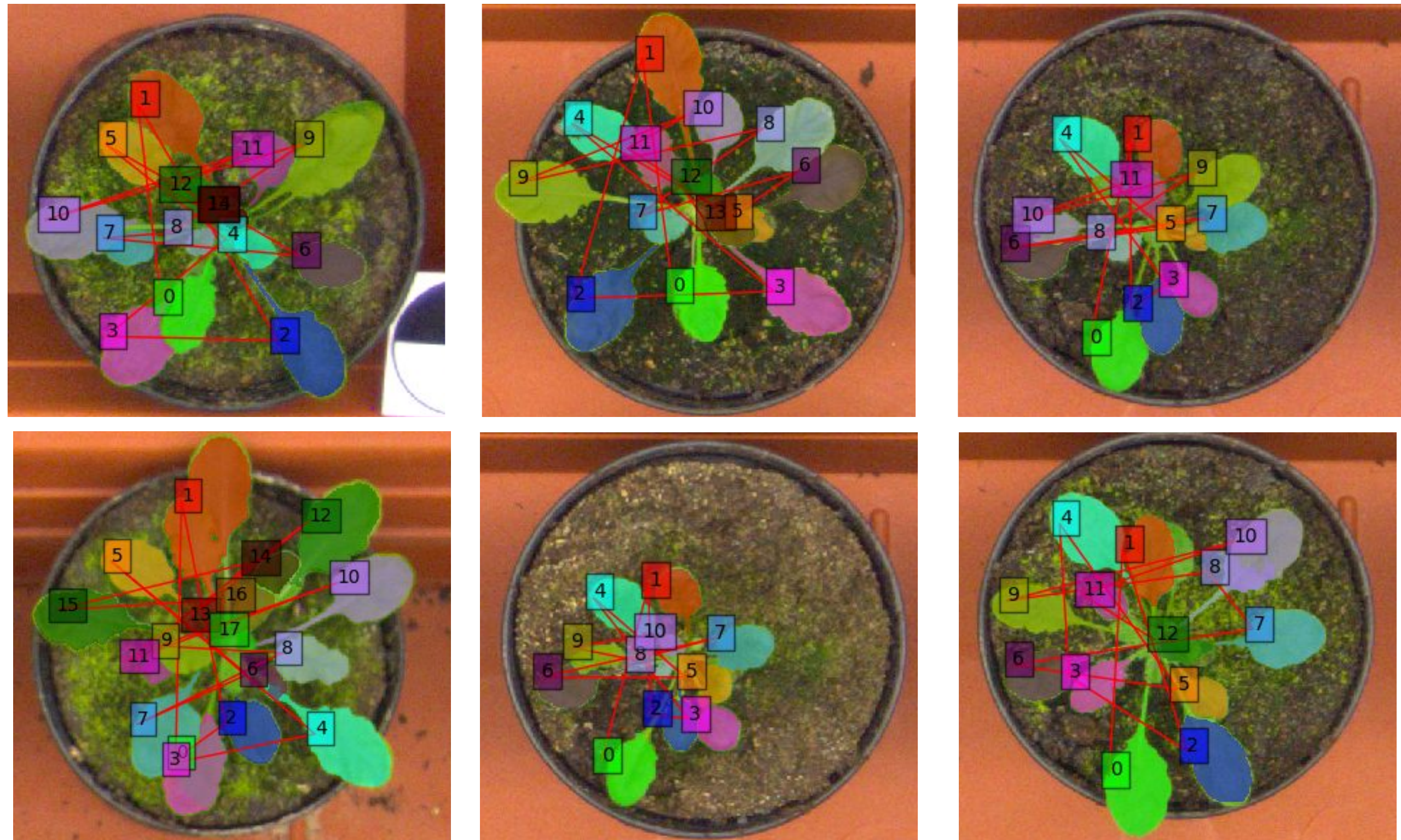}
  \caption{CVPPP}
  \label{fig:vis_leaves}
\end{subfigure}

\begin{subfigure}[t]{\textwidth}
  \centering
  \includegraphics[width=\columnwidth]{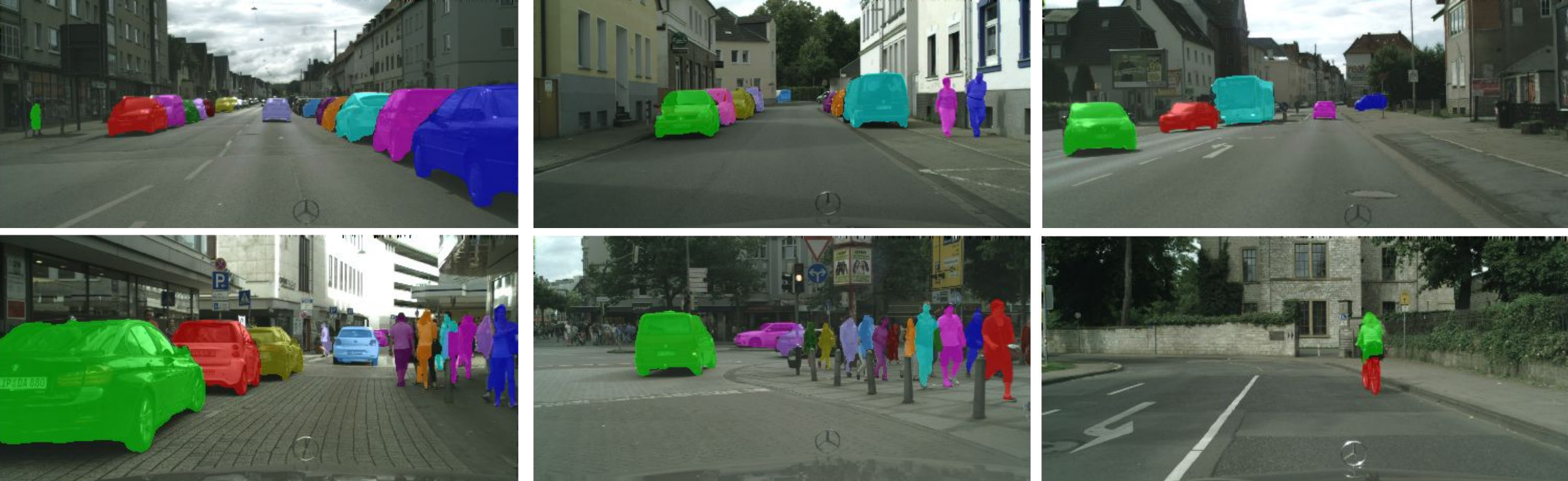}
  \caption{Cityscapes}
  \label{fig:viz_citys}
  \end{subfigure}
  \caption{Examples of generated output sequences for the three datasets.}
  \label{fig:viz_all}
\end{figure}

We first train and evaluate our model with the Pascal VOC dataset. In Table \ref{tab:seq_sota} we compare our method with the recurrent model in \cite{romeraparedes}, whose approach is the most similar to ours. However, since they train and evaluate their method on the person category only, we report the results for this category separately despite that our model is trained for all 20 categories. We outperform their results by a significant margin ($AP_{50}$ of $46.6$ vs. $60.7$), even in the case in which they use a post processing based on CRFs, reaching an $AP_{50}$ of $50.1$.  Figure \ref{fig:pascal_examples} shows examples of predicted object sequences for Pascal VOC images.
Table \ref{ttab:voc_thresh} compares our approach with non-sequential methods.
We outperform early proposal-based ones \cite{sds,multi_instance} by a significant margin across all IoU thresholds. Compared to more recent works \cite{r2ios,arnab_bmvc,pfn,pixelwise}, our method falls behind for lower thresholds, but remains competitive and even superior in some cases for higher thresholds.



In the case of the CVPPP dataset, our method also outperforms the one in \cite{romeraparedes} by a significant margin. However, the sequential model in \cite{renzemel} obtains better results in this benchmark. Their method incorporates an input pre-processing stage and involves multi-stage training with different levels of supervision. In contrast with \cite{renzemel}, our method directly predicts binary masks from image pixels without imposing any constraints regarding the intermediate feature representation. In Figure \ref{fig:vis_leaves} we show examples of predictions obtained by our model for this dataset. Although the number of objects is much higher in this benchmark than in Pascal VOC, our model is able to accurately output one object at a time.





Our performance on Cityscapes is comparable to the results of the only sequential method previously evaluated on this dataset \cite{renzemel}, but does not meet state of the art results obtained by non-sequential methods, which reach $AP_{50}$ figures of 58.1 \cite{maskrcnn}, 35.9 \cite{vangool} and 35.3 \cite{watershed_urtasun}.
Figure \ref{fig:viz_citys} depicts some sample predictions of our model for this dataset. 
While our approach is competitive or even better than \cite{renzemel} for simpler and frequent objects (e.g.~$AP_{50}$ figures of $45.7$ vs. $41.9$ for \textit{car}, and $20.5$ vs. $21.2$ for \textit{person}), it obtains lower scores for less frequent and commonly smaller instances (e.g.~$2.8$ vs. $10.5$ for \textit{bike} and $6.8$ vs. $14.7$ for \textit{motorbike})\footnote{Detailed metrics for all categories are reported in the supplementary material.}. We hypothesize that, as the segmentation module in \cite{renzemel} extracts features at a local scale once the detection module predicts a bounding box, their model can accurately predict binary masks for small instances. In contrast, our method operates at global scale for all instances, generating one binary mask at a time considering all pixels in the image. Working with images at higher resolution would allow us to improve our metrics (specially for small objects), which would come at a cost of higher computational requirements. 
It is also worth noting that the classification scores in \cite{renzemel} are provided by a separate module trained for the task of semantic segmentation, while our method predicts them together with the binary masks. To the best of our knowledge, ours is the first recurrent model used as a solution for Cityscapes.

\subsection{Ablation studies}

In this section, we quantify the effect of each of the components in our network (encoder, skip-connections and number of recurrent layers). Table \ref{ttab:ablation} presents the results of these experiments for Pascal VOC. First, we compare the performance of different image encoders. We find that a deeper encoder yields better performance, with a $23.87\%$ relative increase from VGG-16 to ResNet-101. Further, we analyze the effect of using different skip connection modes (i.e.~summation, concatenation and multiplication), as well as removing them completely. While there is little difference between the different skip connection modes, concatenation has better performance. Completely removing skip connections causes a drop of performance of $6.6\%$, which demonstrates the effectiveness of using them to obtain accurate segmentation masks. We also quantify the effect of reducing the number of \ConvLSTM layers in the decoder. To remove \ConvLSTM layers, we simply truncate the decoder chain and the output of the last \ConvLSTM is upsampled to match the image dimensions. This becomes the input to the last convolutional layer that outputs the final mask. Removing a \ConvLSTM layer also means removing the corresponding skip connection. (e.g.~if we remove the last \ConvLSTM layer, the features from the first convolutional block in the encoder are never used in the decoder). Results in table \ref{ttab:ablation} show a decrease in performance as we remove layers from the decoder, which indicates that both the depth of the decoder and the skip connections coming from the encoder contribute to the result. Notably, keeping the original five \ConvLSTM layers in the decoder but removing the skip connections provides a similar performance as using a single \ConvLSTM layer without skip-connections (AP of $53.3$ against $53.2$). This indicates that a deeper recurrent module can only improve performance if the side outputs from the encoder are used as additional inputs.

\begin{table}
\begin{subtable}[t]{.5\textwidth}
\centering
\resizebox{\textwidth}{!}{
\begin{tabular}[b]{@{}ccccc@{}}
\toprule
\textbf{Encoder} & \textbf{skip}   & $\mathbf{N}$ & $\mathbf{AP_{50}}$ & \textbf{$\mathbf{AP_{person,50}}$} \\ \midrule
VGG16   & concat & $5$   & $46.5$  & $51.7$         \\
R50     & concat & $5$   & $53.0$  & $53.9$          \\
R101    & concat & $5$   & $\mathbf{57.0}$  & $\mathbf{60.7}$         \\ \midrule
R101    & sum    & $5$   & $56.7$  & $57.8 $         \\
R101    & mult   & $5$   & $56.1$  & $59.2$          \\
R101    & none   & $5$   & $53.8$  & $51.3$          \\ \midrule
R101    & concat & $4$   & $56.0$  & $59.0$        \\
R101    & concat & $3$   & $56.1$  & $59.5$        \\
R101    & concat & $2$   & $54.5$  & $54.0$         \\
R101    & -      & $1$   & $53.3$  & $50.6$         \\ \bottomrule
\end{tabular}
}
\subcaption{}
\label{ttab:ablation}

\end{subtable}  
\begin{subtable}[t]{.5\textwidth}
\centering
\resizebox{\textwidth}{!}{
\begin{tabular}[b]{@{}lcccc@{}}
\toprule
\textbf{Model}    & $\mathbf{AP_{50}}$   & $\mathbf{AP_{60}}$  & $\mathbf{AP_{70}}$  & $\mathbf{AP_{80}}$  \\ 
\midrule
SDS \cite{sds}                     & $43.8$  & $34.5$ & $21.3$ & $8.7$                             \\
Chen et al. \cite{multi_instance}  & $46.3$  & $38.2$ & $27.0$   & $13.5$                            \\
PFN \cite{pfn}                     & $58.7$  & $51.3$ & $42.5$ & $31.2$                            \\
R2-IOS \cite{r2ios}                & $\mathbf{66.7}$  & $\mathbf{58.1}$ & $46.2$ & $-$                                \\
Arnab et al. \cite{arnab_bmvc}    & $58.3$  & $52.4$ & $45.4$ & $34.9$                          \\
Arnab et al. \cite{pixelwise}      & $61.7$  & $55.5$ & $\mathbf{48.6}$ & $\mathbf{39.5}$                              \\
MPA \cite{mpa}                     & $60.3$  & $54.6$ & $45.9$ & $34.3$                              \\
Ours                   & $57.0$  & $51.8$ & $41.5$ & $37.8$                             \\ 
\bottomrule
\end{tabular}
}
\subcaption{}
\label{ttab:voc_thresh}
  \end{subtable}  
\label{ttab:pascal_voc}
\caption{Results for Pascal VOC 2012 validation set. \textbf{(a)} Ablation studies. \textbf{(b)} Comparison with the state of the art for different IoU thresholds.}
\end{table}



\subsection{Error analysis}

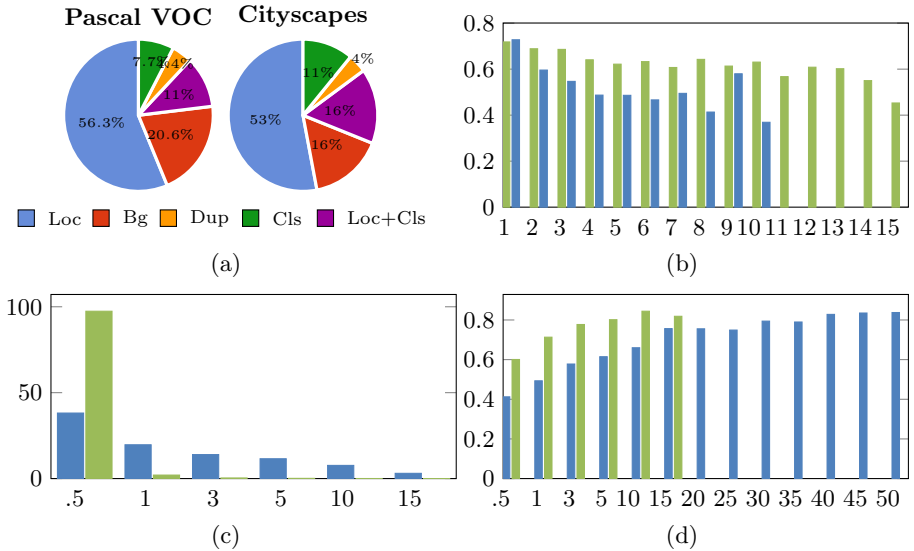
\begin {figure*}[t!]
\begin{subfigure}[t]{0.5\textwidth}
        \centering
\begin{tikzpicture}
[
    pie chart,
    slice type={comet}{blu},
    slice type={legno}{rosso},
    slice type={coltello}{giallo},
    slice type={sedia}{viola},
    slice type={caffe}{verde},
    pie values/.style={font={\tiny}},
    scale=1
]

    \pie{Pascal VOC}{56.3/comet,20.6/legno,11/sedia,4.4/coltello,7.7/caffe}
    \pie[xshift=2.2cm,values of coltello/.style={pos=1.1}]%
        {Cityscapes}{53/comet,16/legno,16/sedia,4/coltello,11/caffe}
   
    \legend[shift={(-1.5cm,-1cm)}]{{Loc}/comet}
    \legend[shift={(-0.5cm,-1cm)}]{{Bg}/legno}
    \legend[shift={(0.4cm,-1cm)}]{{Dup}/coltello}
    \legend[shift={(1.5cm,-1cm)}]{{Cls}/caffe}
    \legend[shift={(2.5cm,-1cm)}]{{Loc+Cls}/sedia}
\end{tikzpicture}
\caption{}
\label{fig:pie}
\end{subfigure}%
\begin{subfigure}[t]{0.5\textwidth}
        \centering
\begin{tikzpicture}
    \begin{axis}[
        width  =7cm,
        height = 4cm,
        major x tick style = transparent,
        ybar=0.3mm,
        bar width= 1.0mm,
        symbolic x coords={1,2,3,4,5,6,7,8,9,10,11,12,13,14,15},
        x tick label style={rotate=0, anchor=north east, inner sep=0mm},
        xtick = data,
        enlarge x limits=0.02,
        ymin=0,
       legend style={ anchor=north east,legend columns=-1}
    ]

    \addplot[style={ggreen,fill=ggreen,mark=none}]
             coordinates {
(1,0.7183)
(2,0.6886)
(3,0.6859)
(4,0.6406)
(5,0.6217)
(6,0.6327)
(7,0.6072)
(8,0.6425)
(9,0.6133)
(10,0.63064)
(11,0.56740)
(12,0.60842)
(13,0.60182)
(14,0.55004)
(15,0.45291)};

        \addplot[style={bblue,fill=bblue,mark=none}]
            coordinates {
(1,0.7277)
(2,0.5960)
(3,0.5468)
(4,0.4871)
(5,0.4862)
(6,0.4663)
(7,0.4947)
(8,0.4134)
(9,0.5798)
(10,0.3690)
};

    \end{axis}

\end{tikzpicture}
	\caption{}
    \label{fig:iou_tstep}
\end{subfigure}%

\begin{subfigure}[t]{0.5\textwidth}
        \centering
\begin{tikzpicture}
    \begin{axis}[
        width  =7cm,
        height = 4cm,
        major x tick style = transparent,
        ybar=0.3mm,
        bar width= 3.5mm,
        symbolic x coords={.5,1,3,5,10,15},
        x tick label style={rotate=0, anchor=north east, inner sep=0mm},
        xtick = data,
        enlarge x limits=0.1,
        ymin=0,
       legend style={ anchor=north east,legend columns=-1}
    ]
        \addplot[style={bblue,fill=bblue,mark=none}]
            coordinates {
(.5,38.1420)
(1,19.6769)
(3,13.9642)
(5,11.5407)
(10,7.6746)
(15,3.0583)

};

        \addplot[style={ggreen,fill=ggreen,mark=none}]
             coordinates {

(.5, 97.4154)
(1,2.0384)
(3, 0.3597)
(5, 0.1332)
(10, 0.0400)
(15, 0.0133)

};
             
    \end{axis}

\end{tikzpicture}
   
	\caption{}
    \label{fig:fn_size}
    
   \end{subfigure}%
 \begin{subfigure}[t]{0.5\textwidth}
 \centering
\begin{tikzpicture}
    \begin{axis}[
        width  =7cm,
        height = 4cm,
        major x tick style = transparent,
        ybar=0.3mm,
        bar width= 1.0mm,
        symbolic x coords={.5,1,3,5,10,15,20,25,30,35,40,45,50},
        x tick label style={rotate=0, anchor=north east, inner sep=0mm},
        xtick = data,
        enlarge x limits=0.02,
        ymin=0,
       legend style={ anchor=north east,legend columns=-1}
    ]
        \addplot[style={bblue,fill=bblue,mark=none}]
            coordinates {

(.5,0.4123)
(1,0.4931)
(3,0.5778)
(5,0.6147)
(10,0.6599)
(15,0.7564)
(20,0.7554)
(25,0.7491)
(30,0.7942)
(35,0.7898)
(40,0.8282)
(45,0.8351)
(50,0.8376)

};

        \addplot[style={ggreen,fill=ggreen,mark=none}]
             coordinates {
(.5,0.6007)
(1,0.7132)
(3,0.7772)
(5,0.8013)
(10,0.8442)
(15,0.8186)

};

    \end{axis}

\end{tikzpicture}
   
	\caption{}
    \label{fig:iou_size}
      \end{subfigure}
	\caption{\textbf{(a)} False positive distribution. \textbf{(b-d)} Error analysis on Pascal VOC (blue) and Cityscapes (green): \textbf{(b)} IoU vs time step, \textbf{(c)} False negative size distribution, \textbf{(d)} IoU vs object size (object size given as the image \% it covers). Reported values in \textbf{(a)} and \textbf{(d)} are constrained to the particularities of each dataset (object sequences for Pascal VOC are shorter and objects in Cityscapes are smaller).}
    \label{fig:errors}
\end{figure*}

Following standard error diagnosis studies for object detectors \cite{hoiem2012diagnosing}, we show the distribution of false positive (FP) errors, considering the following types: localization errors (Loc), confusions with the background (Bg), duplicates (Dup), miss-classifications (Cls), and double localization and classification errors (Loc+Cls). Figure \ref{fig:pie} shows that most FPs are caused by inaccurate localization. Further, in Figure \ref{fig:iou_tstep} we show the mask quality in terms of IoU depending on the time step when it was predicted. It can be observed that the quality of the masks degrades as the number of time steps increases. We believe that, as features extracted from the encoder are fixed for any output sequence length, more information has to be encoded in the same feature size for long sequences, acting as a bottleneck. The same applies to the decoder, that must retain more information for longer sequences in order to decide what to output next. These intrinsic properties of a recurrent model may lead to poor mask localization for the last masks of the output prediction. A performance drop for longer sequences when using RNNs has already been demonstrated in other works \cite{bahdanau2014neural}. Further, we analyze the distribution of false negatives in terms of their size with respect to the image dimensions. We cluster objects in different bins according to the image percentage they cover. Figure \ref{fig:fn_size} shows that, for both datasets, most of the false negatives (97\% and 38\% for Cityscapes and Pascal VOC, respectively) are small objects that cover less than 1\% of the image. Figure \ref{fig:iou_size} shows the average IoU for objects of different sizes. Both figures indicate that our method achieves higher IoU values for big objects and struggles with small ones.

\subsection{Object Sorting Patterns}


We observe that the outputs of the model follow a consistent order across images in CVPPP, as depicted in Figure \ref{fig:vis_leaves}. The complexity and scale of Pascal VOC and Cityscapes make this qualitative analysis unfeasible, so we analyze the sorting patterns learned by the network by computing their correlation with three predefined sorting strategies: right to left (\textit{r2l}), bottom to top (\textit{b2t}) and large to small (\textit{l2s}).
We take the center of mass of each object to represent its location and its area as the measure for its size.

We sort the sequence of predicted masks according to one of the strategies and compare the resulting permutation indices with the original ones using the Kendall tau correlation metric: $\tau = \frac{P-Q}{N(N-1)/2}$. Given a sequence of masks $x \in (x_{1},...,x_{N})$ and its permutation $y\in (y_{1},...,y_{N})$, $P$ is the number of concordant pairs (i.e.~pairs that appear in the same order in the two lists) and $Q$ is the number of discordant pairs. $\tau \in [-1,1]$, where 1 indicates complete correlation, -1 inverse correlation and 0 means there is no correlation between sequences. Table \ref{tab:corr_arbitrary} presents the results for this experiment. For simplicity, we do not show the results for the opposite sorting criteria in the table (i.e.~left to right, small to large and top to bottom), since their $\tau$ value would be the same but with the opposite sign. We observe strong correlation with a horizontal sorting strategy for both datasets (right to left in Pascal VOC and left to right in Cityscapes), as well as with bottom to top and large to small patterns.

\begin{table}[h]
\begin{subtable}[t]{.4\textwidth}
\centering
\resizebox{0.9\textwidth}{!}{
\begin{tabular}[b]{@{}lccc@{}}
\toprule
       & \textbf{Pascal VOC} & \textbf{Cityscapes}   \\ \midrule
r2l    & \textbf{0.4916}     & \textbf{-0.4428}        \\
b2t    & 0.2788     & 0.2712      \\
l2s    & 0.2739     & 0.1700       \\ \bottomrule
\end{tabular}
}
\caption{}

\label{tab:corr_arbitrary}
\end{subtable}
\begin{subtable}[t]{.60\textwidth}
\centering
\resizebox{0.9\linewidth}{!}{
\begin{tabular}[b]{lcccccc}
\toprule
        & \multicolumn{2}{c}{\textbf{Pascal VOC}} & \multicolumn{2}{c}{\textbf{CVPPP}} & \multicolumn{2}{c}{\textbf{Cityscapes}} \\ 
        & before         & after         & before         & after         & before      & after       \\ 
        \midrule
$f_{4}$ & $-0.048$      & $-0.062$           & $-0.129$     & $0.232$           & $-0.127$         & $-0.162$  \\
$f_{3}$ & $0.014$         & $-0.005$           & $0.032$      & $0.135$           & $0.279$          & $0.194$      \\
$f_{2}$ & $-0.088$        & $-0.125$     & $-0.317$      & $-0.141$       & $-0.111 $        & $0.144$     \\
$f_{1}$ & $0.008$         & $0.286$        & $\textbf{0.184}$      & $\textbf{0.505}$       &  $0.010$          & $0.188$     \\
$f_{0}$ & $\mathbf{0.274}$        & $\mathbf{0.634}$        & $-0.054$     & $0.147$        & $\textbf{-0.125}$       &$\mathbf{0.209}$    \\ 
\bottomrule
\end{tabular}
}
\caption{}

\label{tab:conv_corr}
\end{subtable}
\caption{Analysis of object sorting patterns. Correlation values are given by the Kendall tau coefficient $\tau$. \textbf{(a)} Correlation with predefined patterns. \textbf{(b)} Correlation with convolutional activations. $f_{4\ldots0}$ correspond to the output of $\text{ResBlock}_{1\ldots5}$ in ResNet-101, respectively.}  
\label{ttab:sorting_results}

\end{table}

\begin{figure}
\vspace*{-\baselineskip}
  \centering
  \includegraphics[width=\columnwidth]{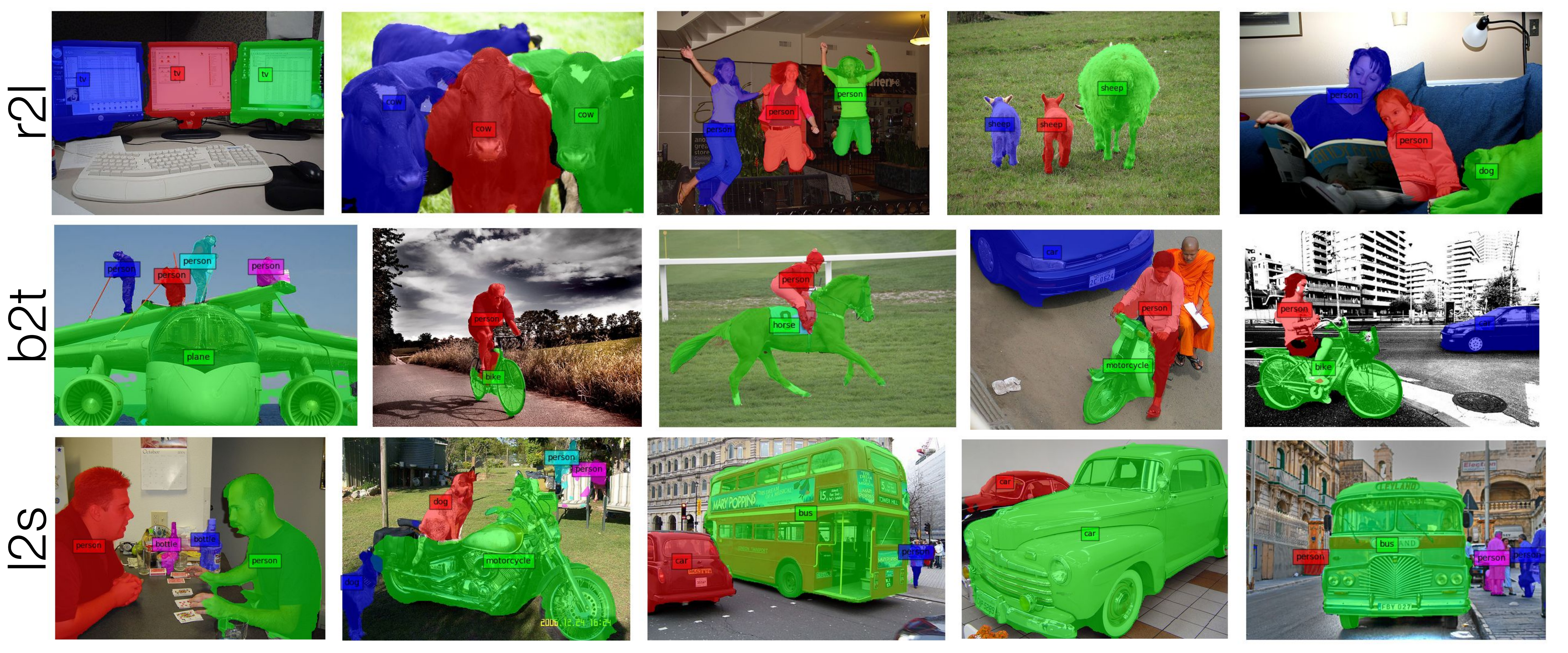}
  \caption{Examples of predicted object sequences for images in Pascal VOC 2012 validation set that highly correlate with the different sorting strategies.}
  \label{fig:viz_pairs}
\end{figure}

Figure \ref{fig:viz_pairs} shows images in Pascal VOC that present high correlation with each of the three sorting strategies. Interestingly, the model adapts its scanning pattern based on the image contents, choosing to start from one side when objects are next to each other, or starting from the largest one when the remaining objects are much smaller. 
The pattern in Cityscapes is more consistent,
which we attribute to the similar structure present in all the images in the dataset. First, the objects in both sides of an image are predicted, starting with the left side; then the model segments the objects in the middle while following similar patterns to the ones in Pascal VOC. This pattern can be observed in Figure \ref{fig:viz_citys}.

\begin {figure}
\vspace*{-\baselineskip}
\centering
\begin{adjustbox}{width=\columnwidth}
\begin{tikzpicture}
    \begin{axis}[
        width  =15cm,
        height = 5.5cm,
        major x tick style = transparent,
        ybar=0.3mm,
        bar width= 1.0mm,
        symbolic x coords={bird+bird, tv+tv, boat+boat, horse+horse, car+car, chair+chair, person+person, motor+motor, cow+cow, sheep+sheep,person+bottle, table+person, person+car, person+dog, motor+person, person+chair, horse+person, table+chair, bicycle+person},
        x tick label style={rotate=45, anchor=north east, inner sep=0mm},
        xtick = data,
        enlarge x limits=0.02,
        ymin=0,
       legend style={ anchor=north east,legend columns=-1}
    ]
        \addplot[style={bblue,fill=bblue,mark=none}]
            coordinates {(bird+bird,0.6000)
(tv+tv,0.6923)
(boat+boat,0.7037)
(horse+horse,0.7083)
(car+car,0.6182)
(chair+chair,0.6094)
(person+person,0.5769)
(motor+motor,0.8182)
(cow+cow,0.9000)
(sheep+sheep,0.5818)
(person+bottle,0.4286)
(table+person,0.7222)
(person+car,0.5238)
(person+dog,0.6471)
(motor+person,0.2368)
(person+chair,0.6667)
(horse+person,0.3714)
(table+chair,0.7143)
(bicycle+person,0.3077)};

        \addplot[style={ggreen,fill=ggreen,mark=none}]
             coordinates {(bird+bird,0.2400)
(tv+tv,0.0769)
(boat+boat,0.1852)
(horse+horse,0.0417)
(car+car,0.1818)
(chair+chair,0.2500)
(person+person,0.1692)
(motor+motor,0.0909)
(cow+cow,0.1000)
(sheep+sheep,0.2909)
(person+bottle,0.2857)
(table+person,0.5556)
(person+car,0.1905)
(person+dog,0.4118)
(motor+person,0.6842)
(person+chair,0.3333)
(horse+person,0.6000)
(table+chair,0.1429)
(bicycle+person,0.7308)
};
             
         \addplot[style={orange,fill=orange,mark=none}]
             coordinates {(bird+bird,0.4400)
(tv+tv,0.3846)
(boat+boat,0.5556)
(horse+horse,0.4583)
(car+car,0.5818)
(chair+chair,0.5000)
(person+person,0.4962)
(motor+motor,0.3636)
(cow+cow,0.4667)
(sheep+sheep,0.4545)
(person+bottle,0.6429)
(table+person,0.6667)
(person+car,0.4762)
(person+dog,0.7059)
(motor+person,0.8158)
(person+chair,0.5714)
(horse+person,0.7429)
(table+chair,0.5000)
(bicycle+person,0.5000)};

        \legend{r2l,b2t,l2s}
    \end{axis}

\end{tikzpicture}
\end{adjustbox}
   
	\caption{Percentage of consecutive object pairs of different categories that follow a particular sorting pattern.}
    \label{fig:pairs}
\end{figure}
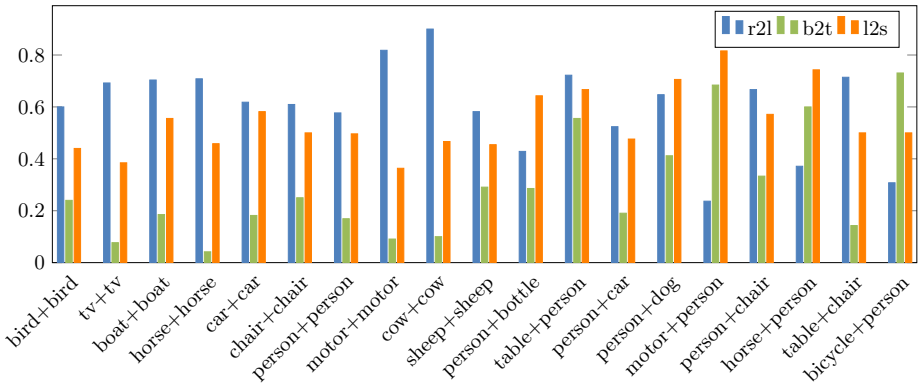

\vspace*{-\baselineskip}
Further, we quantify the number of object pairs in Pascal VOC images that are predicted in each of the predefined orders. For a pair of objects $o_{1}$ and $o_{2}$ that are predicted consecutively, we can say they are sorted in a particular order if their difference in the axis of interest is greater than 15\% (e.g.~a pair of consecutive objects follows a right to left pattern if the second object is to the left of the first by more than $0.15 W$ pixels, being $W$ the image width). Figure \ref{fig:pairs} shows the results for object pairs separated by category. For clarity, only pairs of objects that are predicted together at least 20 times are displayed. We observe a substantial difference between pairs of instances from the same category and pairs of objects of different classes. While same-class pairs seem to be consistently predicted following a horizontal pattern (right to left), pairs of objects from different categories are found following other patterns reflecting the relationships between them. For example, the pairs $motorcycle+person$, $bicycle+person$ or $horse+person$ are often predicted following the vertical axis, from the bottom to the top of the image, which is coherent with the usual spatial distribution of objects of these categories in Pascal VOC images.

\begin{figure}
  \vspace{-\baselineskip}
  \centering
  \includegraphics[width=\columnwidth]{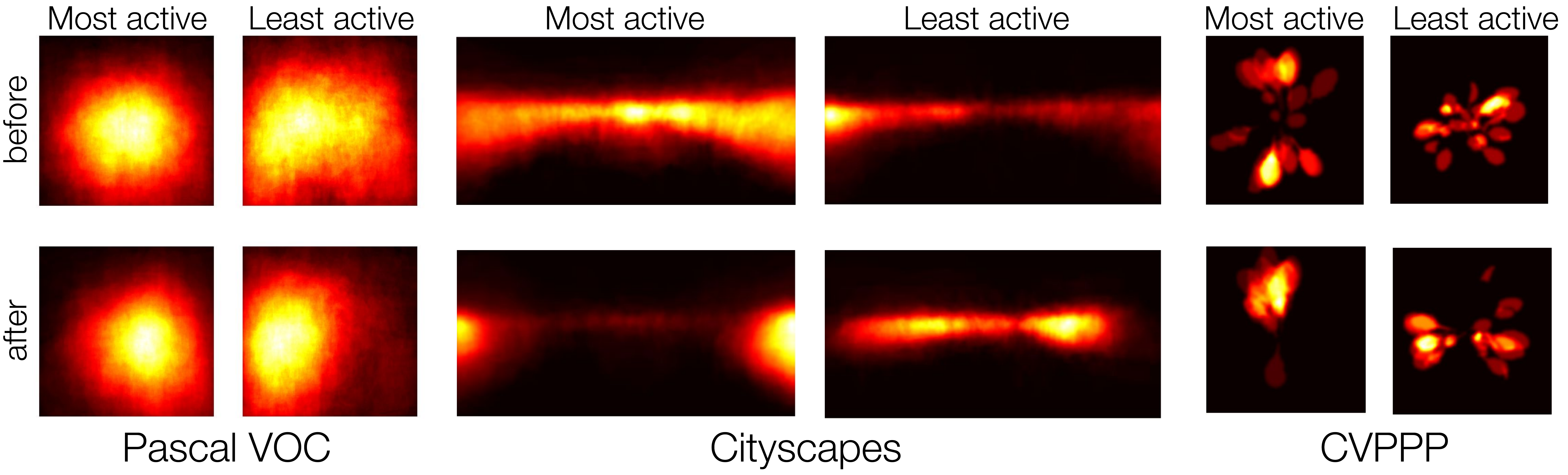}
  \caption{Most and least active objects in last (Pascal VOC and Cityscapes) and second to last (CVPPP) block in the encoder before and after training.}
  \vspace{-\baselineskip}
  \vspace*{-\baselineskip}
  \label{fig:first_object_b_a}
\end{figure}

\vspace*{-\baselineskip}
We also check whether the order of the predicted object sequences correlates with the features from the encoder. Since these are the inputs to the recurrent layers in the decoder (which do not change across different time steps), the network must learn to encode the information of the object order in these activations. To test whether this is true, we permute the object sequence based on the activations in each of the convolutional layers in the encoder and check the correlation with the original sequence. Table \ref{tab:conv_corr} shows the Kendall tau correlation values of predicted sequences with these activations, before and after training the model. We observe that correlation increases after training the model for our task. The predicted sequences correlate the most with the activations in the last block in the encoder both for Pascal VOC and Cityscapes. This is a reasonable behavior, since those features are the input to the first \ConvLSTM layer in the decoder. In the case of images from the CVPPP dataset, we find that the predicted object sequences correlate with the activations in the second to last convolutional layer in the encoder. We hypothesize that the semantics in the last layer of the encoder, which is pretrained on ImageNet, are not as informative for this task. In Figure \ref{fig:first_object_b_a} we display the most and least active object in the most correlated block in the encoder for each dataset. We show figures for features before and after training the model. For Pascal VOC images, we observe a shift of the most active objects from the center of the image to the bottom-right part of the image, while the least active objects are located in the left part of the image. In the case of Cityscapes, the most active objects move from the center to right-most and left-most part of the image after training. Regarding CVPPP, we observe that the network learns a specific route to predict leaves which is consistent across different images, starting in the top-most part of the image. 

\section{Conclusion}
\label{sec:conclusion}

We have presented a recurrent method for end-to-end semantic instance segmentation, which can naturally handle variable length outputs by construction. Unlike proposal-based methods, which generate an excessive number of predictions and rely on an external post-processing step for filtering them out, our model is able to directly map pixels to the final instance segmentation masks. This allows our model to be optimized for an objective which better matches the conditions of the target task at inference time than those in proposal-based methods. We observed coherent patterns in the order of the predictions that depend on the input image, suggesting that the model makes use of its previous predictions to reason about the next object to be detected. In contrast with other sequential methods that use direct feedback from their output, the choice of a multi-layer recurrent network also has the advantage of being more parallelizable across time steps on modern hardware \cite{cudnn}.



We have detected two main sources of limitations in the proposed model, namely inaccurate masks for small objects and difficulties handling long sequences. The quality of the segmentation for small objects can be improved by increasing the resolution of the input images, although this comes at the cost of a larger memory footprint that can preclude training for long sequences unless the model is parallelized across different GPUs \cite{sutskever2014_seq2seq}. 
In order to improve the performance on long sequences, the memory of the model can be increased by adding more units to each ConvLSTM \cite{collins2016capacity}. There is evidence that the optimal number of units is very dependent on the dataset \cite{alvarez2016learning}, but models with more parameters are also slower to train and require more memory. Finding the best trade-off between performance and computational requirements for each dataset remains as future work.



\section{Acknowledgements}

This work was partially supported by the Spanish Ministry of Economy and Competitivity under contracts TIN2012-34557 by the BSCCNS Severo Ochoa program (SEV-2011-00067), and contracts TEC2013-43935-R and TEC2016-75976-R. It has also been supported by grants 2014-SGR-1051 and 2014-SGR-1421 by the Government of Catalonia, and the European Regional Development Fund (ERDF).
We acknowledge the support of NVIDIA Corporation for the donation of GPUs.

\clearpage

\bibliographystyle{splncs}
\bibliography{egbib}
\end{document}